\documentclass[sigconf]{acmart}
\AtBeginDocument{%
  \providecommand\BibTeX{{%
    \normalfont B\kern-0.5em{\scshape i\kern-0.25em b}\kern-0.8em\TeX}}}

\usepackage{soul}
\usepackage{graphicx}
\usepackage{booktabs}
\usepackage{float}
\usepackage{overpic}
\usepackage{float}  %设置图片浮动位置的宏包
\usepackage{subfloat}
\usepackage{stfloats} % for pic
\usepackage[skip=0.5\baselineskip]{caption}
\usepackage{bm}
\usepackage{amsmath}
\usepackage{booktabs}
\usepackage{multirow}
\usepackage{multicol}
\usepackage{enumitem}
\usepackage{subcaption}
\usepackage{threeparttable}
\usepackage{xspace}
\usepackage{color}
\usepackage[normalem]{ulem}
\usepackage{algorithm}
\usepackage{algorithmic}
\usepackage{marvosym}   % 添加通讯作者的信封
\usepackage{colortbl}   % 表格中的阴影颜色
\usepackage{tcolorbox}  % 文本加框
\tcbuselibrary{breakable}   % 配合文本加框，使得框可以跨页
\usepackage{wasysym}    % 添加实心圆空心圆
\usepackage[figuresright]{rotating} % 旋转表格

\makeatletter
\def\@ACM@checkaffil{% Only warnings
    \if@ACM@instpresent\else
    \ClassWarningNoLine{\@classname}{No institution present for an affiliation}%
    \fi
    \if@ACM@citypresent\else
    \ClassWarningNoLine{\@classname}{No city present for an affiliation}%
    \fi
    \if@ACM@countrypresent\else
        \ClassWarningNoLine{\@classname}{No country present for an affiliation}%
    \fi
}
\makeatother
\usepackage{color, xspace}

\usepackage{xspace}

\newcommand{\name}{NEZHA\xspace}

\copyrightyear{2026}
\acmYear{2026}
\setcopyright{cc}
\setcctype{by-nc-nd}
\acmConference[WWW '26]{Proceedings of the ACM Web Conference 2026}{April 13--17, 2026}{Dubai, United Arab Emirates}
\acmBooktitle{Proceedings of the ACM Web Conference 2026 (WWW '26), April 13--17, 2026, Dubai, United Arab Emirates}
\acmPrice{}
\acmDOI{10.1145/3774904.3792797}
\acmISBN{979-8-4007-2307-0/2026/04}

\title{NEZHA: A Zero-sacrifice and Hyperspeed Decoding Architecture for Generative Recommendations}
\author{Yejing Wang}
\email{yejing.wang@my.cityu.edu.hk}
\affiliation{
    \institution{City University of Hong Kong}
    \city{Hong Kong SAR}
    \country{China}
}
\authornote{Work was conducted during the internship of Yejing Wang at Alibaba.}
\authornote{Yejing Wang and Shengyu Zhou contributed equally to this work..}
\author{Shengyu Zhou}
\authornotemark[2]
\affiliation{
    \institution{Alibaba Group}
    \city{Beijing}
    \country{China}
}

\author{Jinyu Lu}
\affiliation{
    \institution{Alibaba Group}
    \city{Beijing}
    \country{China}
}

\author{Ziwei Liu}
\affiliation{
    \institution{City University of Hong Kong}
    \city{Hong Kong SAR}
    \country{China}
}

\author{Langming Liu}
\affiliation{
    \institution{Alibaba Group}
    \city{Hangzhou}
    \country{China}
}

\author{Maolin Wang}
\affiliation{
    \institution{City University of Hong Kong}
    \city{Hong Kong SAR}
    \country{China}
}
\author{Wenlin Zhang}
\affiliation{
    \institution{City University of Hong Kong}
    \city{Hong Kong SAR}
    \country{China}
}

\author{Feng Li}
\affiliation{
    \institution{Alibaba Group}
    \city{Beijing}
    \country{China}
}

\author{Wenbo Su}
\affiliation{
    \institution{Alibaba Group}
    \city{Beijing}
    \country{China}
}

\author{Pengjie Wang}
\affiliation{
    \institution{Alibaba Group}
    \city{Beijing}
    \country{China}
}

\author{Jian Xu}
\affiliation{
    \institution{Alibaba Group}
    \city{Beijing}
    \country{China}
}
\author{Xiangyu Zhao}
\email{xianzhao@cityu.edu.hk}
\affiliation{
    \institution{City University of Hong Kong}
    \city{Hong Kong SAR}
    \country{China}
}

\authornote{Xiangyu Zhao is the corresponding author.}

\begin{document}
\begin{sloppypar}   % 自动换行和对齐，防止一行文字溢出

% \thanks{$\dagger$ Work was conducted during the internship of Yejing Wang at Alibaba.}
% \thanks{* Xiangyu Zhao is the corresponding author.}
% \affiliation{
% 	\institution{$^1$City University of Hong Kong, $^2$Alibaba Group}
% 	\country{}
% }
%\email{adave631@gmail.com, xianzhao@cityu.edu.hk, tongxu@ustc.edu.cn, kevinxwu@tencent.com}
\renewcommand{\shortauthors}{Yejing Wang, et al.}

%% The abstract
\begin{abstract}
Generative Recommendation (GR), powered by Large Language Models (LLMs), represents a promising new paradigm for industrial recommender systems. However, their practical application is severely hindered by high inference latency, making them infeasible for high-throughput, real-time services and limiting their overall business impact. While Speculative Decoding (SD) has been proposed to accelerate the autoregressive generation process, existing implementations introduce new bottlenecks: they typically require separate draft models and model-based verifiers, which require additional training and increase latency overhead. 
In this paper, we address these challenges with \name, a novel architecture that achieves hyperspeed decoding for GR systems without sacrificing recommendation quality. Specifically, \name integrates a nimble autoregressive draft head directly into the primary model, enabling efficient self-drafting. This design, combined with a specialized input prompt structure, preserves the integrity of sequence-to-sequence generation. Furthermore, to tackle the critical problem of hallucination—a major source of performance degradation—we introduce an efficient, model-free verifier based on a hash set. We demonstrate the effectiveness of \name through extensive experiments on public datasets and have successfully deployed the system on \textbf{Taobao} since October 2025, achieving \textbf{1.2\%} business improvement, translating to \textbf{billion-level advertising revenue} and serving \textbf{hundreds of millions of daily active users}. The code is available at \url{https://github.com/Applied-Machine-Learning-Lab/WWW2026\_NEZHA}.%{this Github page}.
% \zxy{code} \zxy{what platform, time, user number, performance}
\end{abstract}



%% CCS Concepts
% <concept_id>10002951.10003317.10003347.10003350</concept_id>
% <concept_desc>Information systems~Recommender systems</concept_desc>

% \ccsdesc[500]{Information systems~Recommender systems}


\maketitle

\section{Introduction}
The extraordinary capabilities of Large Language Models (LLMs) have catalyzed a wide range of applications in recommender systems that leverage LLMs' abilities~\cite{zhu2025rankmixer,zhang2025onetrans,zhai2024actions}, including world knowledge integration~\cite{xi2024KAR,fu2025unified}, representation learning~\cite{ren2024representation}, and semantic understanding~\cite{bao2025bi,zhang2025notellm}. 
Among these LLM-driven solutions, Generative Recommendation (GR)~\cite{zheng2024adapting,ju2025generative,yang2024unifying,gao2025generative} has rapidly emerged as a dominant paradigm due to its outstanding performance and notable strengths in addressing the cold-start problem and enhancing recommendation diversity~\cite{rajput2023recommender,wang2025gflowgr}. Consequently, GR has seen widespread adoption in various industrial applications and production systems~\cite{zhou2025onerec,zheng2025ega,guo2025onesug,wei2025oneloc,chen2025onesearch}.

\begin{table}[t]
\begin{tabular}{@{}c|c|c|cccc@{}}
\toprule
Method & \begin{tabular}[c]{@{}c@{}}Model\\ Size\end{tabular} & \begin{tabular}[c]{@{}c@{}}Beam\\ Size\end{tabular} & Prefill & Decode & System & Total \\ \midrule
Vanilla & 3B   & \multirow{3}{*}{512} & 1.58 & 6.87 & 0.96 & 9.41 \\ % 0.73
Vanilla & 0.6B &                      & 1.0  & 2.95 & 0.91 & 4.86 \\ % 60%  
NEZHA   & 0.6B &                      & 1.0  & 0.78 & 0.08 & 1.86 \\ \bottomrule % 0.42
\end{tabular}
\caption{Serving latency decomposition.}\label{tab:latdec}
\vspace{-6mm}
\end{table}

% \begin{figure}[t]
% \centering
% % \vspace{-1mm}
% \includegraphics[width=1.04 \linewidth]{fig/Intro.pdf}
% \caption{
% Overview of the GR Deployment Pipeline using the Qwen model as an example. 
% \textbf{(a) Item Tokenization: }An item from the catalog, such as a ``phone case for iPhone 17'', is first tokenized into a structured, multi-token semantic ID based on its multi-modal representation.
% \textbf{(b) LLM Training:} The LLM is then fine-tuned on a large offline dataset ($\mathcal{D}$) of user interactions to learn the patterns for predicting the next item's semantic ID based on user context.
% \textbf{(c) LLM Serving:} Finally, the trained model is deployed for real-time serving. Given an incoming user request, it autoregressively generates the semantic ID of the recommended item.
% }
% \label{fig:intro}
% % \vspace{-9mm}
% \end{figure}

% As illustrated in Figure~\ref{fig:intro}, 
The typical GR deployment involves three stages~\cite{li2025survey}: (a) item tokenization~\cite{hua2023index,wang2024learnable,zheng2025universal}, (b) LLM training~\cite{deng2025onerec,yang2025sparse}, and (c) LLM serving~
\cite{xi2025efficiency,lin2025efficient,xi2024decoding}. While the first two stages are performed offline, the serving latency of the final stage presents a formidable obstacle to large-scale industrial adoption, particularly in latency-sensitive scenarios~\cite{wang2025put}. For instance, in Taobao's search advertising, a business unit that accounts for over a quarter of the company's revenue, the core service requires a response time of less than 30 milliseconds. However, the latency of the current GR solution we
deployed in this scenario exceeds 1 second, thus hindering the full commercial potential of real-time serving.

% \usepackage{booktabs}
% \usepackage{multirow}

% As illustrated in Figure~\ref{fig:frame}, the LLM inference process includes two steps: LLM prefilling and decoding. The prefilling refers to the encoding of the input while the decoding refers to the generation process, e.g., widely-used beam search. The root of the exceeding inference latency lies in the heavy decoding process, as pointed by existing literatures~\cite{xi2025efficiency,leviathan2023fast}. For example, \citet{lin2025efficient} suggest the decoding process can cost nearly ninety percentage of the inference time. To validate this conclusion, we test the serving latency and decompose it into prefilling, decoding,  and system cost. The result is presented in Table~\ref{tab:latdec}, where the cost time is represented by the unit time for the prefilling with LLM-0.6B (we desensitize the specific time and LLM backbone). We present the serving latency for two sizes of LLM (0.6B and 3B) with a beam size as 512 (a common size for recall in practice). We can observe a similar result to the literature~\cite{lin2025efficient} that the decoding conquers more than 60\% of the inference time (prefilling and decoding) from the first two rows. In addition, while the prefilling and system remains the similar when compressing the model size. These observations highlights the urgent need of accelerating the decoding process. 

LLM inference proceeds in two stages: a \textit{prefill stage}, where the LLM processes the input prompt in parallel, followed by a \textit{decoding stage} that autoregressively generates output tokens. For GR, this decoding is typically performed using beam search~\cite{freitag2017beam} to produce a diverse set of high-quality candidate items, leading to the excessive inference latency as identified in prior literature~\cite{xi2025efficiency,leviathan2023fast}. For instance, \citet{lin2025efficient} reports that the decoding can consume nearly 90\% of the total inference time for Llama-7B.
To empirically validate this claim, we profiled the serving latency and decomposed it into three components: prefilling, decoding, and system overhead. The results are presented in Table~\ref{tab:latdec}, where all timings are normalized by the prefilling time of the 0.6B model for confidentiality. We benchmarked two LLMs of different sizes (0.6B and 3B) using a beam size of 512, a common setting for recall tasks in industrial practice. The first two rows reveal that decoding consistently accounts for over 60\% of the total inference time (prefilling + decoding), corroborating the findings in~\cite{lin2025efficient}. Furthermore, we observe that while model compression significantly reduces the decoding time, the prefilling and system overheads remain relatively stable. \textit{These empirical observations underscore the urgent need for a dedicated solution to accelerate the decoding stage.}

Algorithm-level solutions for this problem, i.e., efficient LLM inference, broadly fall into two categories~\cite{wanefficientllm}: KV-cache optimization~\cite{xi2024decoding} and speculative decoding (SD)~\cite{lin2025efficient,zagyva2025speed}. However, optimizing the KV cache alone is often insufficient: these techniques, e.g., FlashAttention~\cite{dao2022flashattention}, yield negligible efficiency gains when generating the extremely short sequences (e.g., 3-4 tokens) common in GR~\cite{daoflashattention}. The results presented in Table~\ref{tab:latdec}, which already incorporate KV-cache enhancements, still significantly exceed the requirements of industrial applications. Consequently, this paper focuses on developing SD solutions to bridge the efficiency gap. Despite considerable efforts for efficient LLM inference~\cite{zhang2025surveyparallel,xia2024unlocking,zhang2024beyond}, a deployable solution for industrial-scale GR systems remains elusive. Existing SD frameworks tailored for applications exhibit critical shortcomings in both core stages: drafting and verification~\cite{zhang2024beyond,zhang2025surveyparallel}:
\begin{itemize}[leftmargin=*]
    \item \textbf{Drafting}: Current approaches predominantly rely on external draft models, such as smaller language models~\cite{lin2025efficient,zagyva2025speed} or retrieval models~\cite{ding2024inductive,xi2025efficiency}. This strategy not only necessitates the costly maintenance of an extra model but also introduces additional inference overhead, offsetting the intended speed gains.
    \item \textbf{Verification}: Prevailing methods still depend on invoking the original large model to validate the drafted tokens~\cite{zagyva2025speed,lin2025efficient,xi2025efficiency,ding2024inductive}. This reliance means they fail to completely avoid the time-consuming decoding steps of LLM, thereby placing a fundamental ceiling on the achievable acceleration.
\end{itemize}

% Despite the considerable efforts on designing SD frameworks for efficient LLM~\cite{zhang2025surveyparallel,xia2024unlocking,zhang2024beyond}, there still lack a deployable SD solution for industrial GR systems. Existing frameworks on this topic, SD applications, fall short on both steps of SD, i.e., \textbf{drafting} and \textbf{verification}. Specifically, they rely on external models for drafting, either small language models~\cite{lin2025efficient,zagyva2025speed} or retrieval models~\cite{ding2024inductive,xi2025efficiency}. This increases the training of an extra draft model and introduces additional inference latency for serving. As for the verification, existing methods typically verify the draft model predictions with the original large model~\cite{zagyva2025speed,lin2025efficient,xi2025efficiency,ding2024inductive}, failing to totally dropping the time-consuming decoding steps, limiting the acceleration effect. 

For these reasons, we design the \textbf{N}imble drafting and \textbf{E}fficient verification, thus propose a \textbf{Z}ero-sacrifice and \textbf{H}yperspeed decoding \textbf{A}rchitecture for industrial-scale GR systems (\name). Specifically, \name possesses a self-drafting capability~\cite{xia2024unlocking,li2024eagle,cai2024medusa}. This allows the model to efficiently predict the next item by itself, thus avoiding the training and serving of extra models. Additionally, we design a model-free verification method specific to GR problems: we attribute the poor performance of drafted results to hallucination (invalid semantic IDs) and verify the draft item using a hash set. We further boost performance by incorporating the autoregressive drafting head~\cite{li2024eagle,cheng2024recurrent} and setting prompts with special tokens to enforce the sequential integrity required for GR. The proposed \name achieves satisfactory performance and hyperspeed decoding, as outlined in Table~\ref{tab:latdec}, and has been successfully deployed in Taobao, impacting billions-level advertising revenues. We summarize the major contributions of this paper as follows:
\begin{itemize}[leftmargin=*]
    \item  We identify high-latency decoding as the critical barrier to the large-scale deployment of GR systems. To address this, we propose \name, an accelerated architecture designed to meet strict industrial latency requirements.
    % We identify the stumbling block for fully deploying GR framework to industrial systems: the time-consuming decoding. We correspondingly propose \name, accelerating the GR inference to meet the latency threshold. 
    \item We design a novel framework combining a nimble self-drafting mechanism with an efficient, model-free verifier. This approach preserves the high-fidelity output of autoregressive models while effectively mitigating hallucinations, thus ensuring the zero-sacrifice recommendation quality.
    % We design the nimble self-draft component and efficient verifier for \name, maintaining the autoregressive prediction and eliminating the hallucination problem for satisfactory performance. 
    \item We conduct extensive experiments on three public datasets across two LLM backbones. The results demonstrate the superior effectiveness, efficiency, and adaptability of \name.
    % We rigorously examine \name with three public datasets on two item tokenization settings, demonstrating its effectiveness and adaptibility. 
    \item We validate \name's performance on a large-scale industrial dataset and detail its successful deployment in Taobao, a leading e-commerce platform in China. The system now serves \textbf{hundreds of millions of daily active users} and drives \textbf{billion-level increases in advertising revenue}.
    % We provide the evaluation result on industrial dataset and sucesssfully deploy \name in Taobao, the world-leading E-commerce platform, significantly affecting the business. 
\end{itemize}

\noindent\textbf{Significance.} Our work establishes a pivotal advancement in deploying generative recommendation (GR) systems under stringent latency constraints, unlocking their practical viability in real-time, large-scale industrial applications. By effectively addressing the critical challenge of excessive response time that has long hindered GR adoption, \textit{we not only demonstrate a highly efficient serving framework tailored for time-sensitive scenarios but also validate its substantial business impact through successful deployment in production}. This breakthrough redefines the operational boundaries of GR, transforming it from a latency-prohibitive paradigm into a scalable, high-performance solution.

\begin{algorithm}[t]
	\caption{Beam Search for GR Decoding.}\label{alg:beamsearch}
	\raggedright
	{\bf Input}: LLM $\mathcal{M}_p$, query $q$, user sequence $x$, beam size $K$, length of semantic ID $L$\\
	{\bf Output}: $\hat Y_L$ (Top $K$ items represented by semantic IDs)\\
	\begin{algorithmic} [1] 
    \STATE Initialization: $l=1, \hat Y_0=\{\varnothing\}$
        \FOR{$l\leq L$}
            \FOR{$\hat y \in \hat Y_{l-1}$}
    		\STATE Prefilling contexts with LLM: $\boldsymbol{h}_l=\mathcal{M}_p(q,x,\hat y)$
            \STATE Calculating the probabilities of the next token: $\boldsymbol{p}_l=\mathrm{lm\_head}(\boldsymbol{h}_l)$
            \STATE Selecting $K$ tokens with top probabilities: $\{t^k_l, p^k_l\}_{k\leq K}$
            \STATE $\hat Y_l \gets  \hat Y_l \cup [\hat y, t^k_l]  \text{ for } k\leq K$
            \ENDFOR
        \STATE Keeping $K$ candidates in $\hat Y_l$ with the largest cumulative probability ($\prod_{t^k_l\in\hat y}p_l^k, \hat y \in \hat Y_l$)
        \STATE $l\gets l +1$
        \ENDFOR
        \RETURN $\hat Y_L$
	\end{algorithmic}
\end{algorithm}

\section{Preliminary}
This section provides the necessary background on the standard deployment of GR and existing acceleration techniques for autoregressive decoding, with a focus on SD.
\subsection{Generative Recommendation (GR)} 
%is illustrated with the scenario of our E-Commerce Search Advertising in Figure~\ref{fig:intro}, which
The typical process for constructing a GR system consists of three steps: item tokenization, LLM training, and LLM serving:

\noindent\textbf{1. Item Tokenization. } The first stage involves converting each item into multi-token semantic IDs~\cite{rajput2023recommender,STORE,fu2025forge}. In this step, quantization methods (e.g., RQ~\cite{lee2022autoregressive}, PQ~\cite{jegou2010product}) are usually applied to map item contextual representations to discrete codes. These codes are then added to the LLM's vocabulary as special tokens. In this paper, we represent an item $v_i$ as a three-token ID $[t^i_1,t^i_2,t^i_3]$.
%For instance, this paper represents an item $v_i$ as with a three-token ID $[t^i_1,t^i_2,t^i_3]$ based on multi-modal representations (visual and textual), a process visualized in Figure~\ref{fig:intro}(a) with a phone case.

\noindent\textbf{2. LLM Training. } Next, the LLM is trained to understand item semantics and predict user preferences by learning from historical user behavior sequences. The training data consists of tuples $\mathcal{D}=\{(q,x),y\}$, where $q$ is the search query, $x=[v_1,\dots,v_N]$ is the user interaction history, and $y=v_y$ is the ground-truth target item ID (e.g., the item the user clicked). All item IDs are flattened into a sequence of tokens. Specifically, the input token sequence is $[\text{<BOS>},q, t^1_1,t^1_2,t^1_3,\dots,t^N_1,t^N_2,t^N_3,t^y_1,t^y_2,t^y_3,\text{<EOS>}]$ and the loss is only calculated on $[t^y_1,t^y_2,t^y_3]$. %Figure~\ref{fig:intro}(b) visualizes this process with $N=1$.

% (Figure~\ref{fig:intro}(c))
\noindent\textbf{3. LLM Serving. } Finally, the trained model is deployed to generate recommendations for live user requests. For an incoming search query $q$ from a user with history $x$, the model autoregressively generates the semantic ID of the predicted item, token by token, usually with Beam Search~\cite{freitag2017beam}. Denoting the LLM as $\mathcal{M}_p$, this inference process is formally described in Algorithm~\ref{alg:beamsearch}. 
The algorithm unifies the beam search steps by initializing the candidate set with a single empty element  ($\hat Y_0=\{\varnothing\}$ in line 1)). Each step then iterates through the current set of top beams (line 3). For each beam, it generates new, longer sequences by prefilling the context and predicting the next possible tokens (lines 4-7). After all candidates have been expanded, the algorithm prunes the resulting set, retaining only the top $K$ candidates with the highest cumulative probabilities (line 9) for the next iteration or output (line 10\&12). 

A critical bottleneck hinders the full deployment of GR in high-throughput businesses with strict latency requirements. The issue stems from the autoregressive decoding process in beam search, where the LLM $\mathcal{M}_p$ is called iteratively to prefill the context for each new candidate sequence (line 4 in Algorithm~\ref{alg:beamsearch}). Specifically, the LLM is invoked $(K\times(L-1))$ times after the initial context prefill, where $K$ is the beam size and $L$ is the number of tokens in the semantic ID. While $L$ is configurable for different GR systems, $K$ is inevitably large in specific applications; for instance, in item retrieval scenarios, $K$ can be as large as 1,000.
Although optimizations like KV-Caching~\cite{xi2024decoding,dao2022flashattention} can accelerate these LLM calls, they remain insufficient, as evident in Table~\ref{tab:latdec}, which shows that even with KV-Cache, the beam search decoding phase requires nearly triple the time of the initial prefilling for a 0.6B-parameterized LLM. \textit{Consequently, developing novel acceleration techniques is imperative for the industrial adoption of current GR systems.}

\begin{algorithm}[t]
	\caption{Speculative Decoding for GR.}\label{alg:sd}
	\raggedright
	{\bf Input}: target LLM $\mathcal{M}_p$, draft model  $\mathcal{M}_q$, query $q$, user sequence $x$, beam size $K$, length of semantic ID $L$\\
	{\bf Output}: $\hat Y$ (top $K$ items represented by semantic IDs)\\
	\begin{algorithmic} [1] 
    \STATE Initialization: $l=1,\hat Y_0=\{\emptyset\}$
    \WHILE{$l\leq L-1$}
    \STATE $\hat Y_{l},\dots, \hat Y_L \gets \mathrm{BeamSearch}(\mathcal{M}_q,q,x,\hat Y_{l-1},K,L)$
    \STATE Verifying $\hat Y_l,\dots, \hat Y_L$ with $\mathcal{M}_p$ in parallel
    \IF{$\hat Y_m$ is rejected}
    \STATE Updating $\hat Y_m$ with $\mathcal{M}_p$
    \STATE $l\gets m+1$
    \ENDIF
    \ENDWHILE
    \RETURN $\hat Y_L$
	\end{algorithmic}
\end{algorithm}

\subsection{Speculative Decoding (SD)}
Beyond KV-Caching, SD has emerged as the primary choice for algorithm-level acceleration techniques~\cite {xia2024unlocking,zhang2025surveyparallel}. It has garnered significant attention in academia~\cite{lin2025efficient,cai2024medusa,li2024eagle} and recently seen successful deployments in industrial applications~\cite{xi2025efficiency,zagyva2025speed}.

We illustrate the SD for GR in Algorithm~\ref{alg:sd}, where we configure the draft model to generate the entire multi-token semantic ID in a single step. For clarity, we denote the Algorithm~\ref{alg:beamsearch} as $\mathrm{BeamSearch}(\cdot)$ in Algorithm~\ref{alg:sd}, which alternatively predicts future tokens with the draft model $\mathcal{M}_q$ based on the given initialization $\hat Y_{l-1}$ and returns the intermediate candidate set $\hat Y_l$ for any step $l$, not just the final result $\hat Y_L$. The core process involves two main steps: drafting and verification. In the drafting stage, SD employs a smaller, faster draft model $\mathcal{M}_q$ to generate a sequence of candidate tokens for the larger, stronger target LLM $\mathcal{M}_p$ (line 3). The drafted tokens are then verified using a single, parallel forward pass of the target LLM (line 4). Based on this verification, if the draft matches the target model's predictions, it is accepted (line 10). If a mismatch is found, the draft is rejected from the point of the first incorrect token (line 5), and a new token is sampled from the target model's output before the drafting process repeats (line 6-7).

While SD has seen successful industrial applications in domains with relaxed latency requirements, such as travel planning~\cite{zagyva2025speed} and recommendation knowledge generation~\cite{xi2025efficiency}, and has been explored academically for GR~\cite{lin2025efficient}, a production-ready solution for high-throughput businesses remains elusive. These industrial systems operate under severe latency constraints that current SD methods cannot meet. Their drafting process relies on an additional model $\mathcal{M}_q$, while the verification process requires at least one call to the large model $\mathcal{M}_p$. To address these issues, we propose \name, which is based on two key innovations: eliminating the need for an independent draft model through \textbf{self-drafting} and verifying drafts using a \textbf{hash set} without calling $\mathcal{M}_p$.

\section{Methodology}
This section provides a detailed exposition of our proposed method, \name. We begin with a high-level overview of the framework. Subsequently, we examine its two core components: nimble drafting and efficient verification. The section concludes by detailing the training and inference pipeline.

\begin{figure}[t]
\centering
% \vspace{-1mm}
\includegraphics[width=\linewidth]{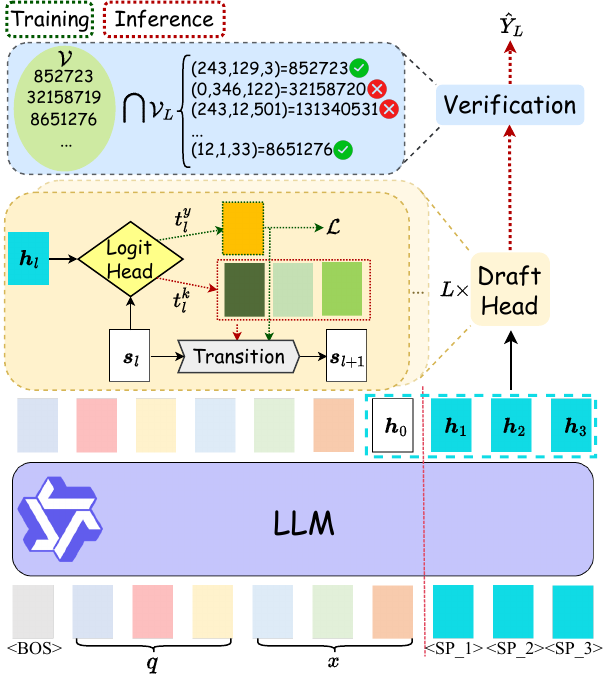}
\caption{An illustration of the \name framework for $L=3,K=3$. The diagram differentiates between the training path (green dotted lines), the inference path (red dotted lines), and the shared computations (black solid lines).}
\label{fig:frame}
% \vspace{-9mm}
\end{figure}

\subsection{Overview}
The \name framework, outlined in Figure~\ref{fig:frame}, introduces several key modifications to the standard decoding process.

First, \name uses a specialized input prompt. Instead of the default format, we append $L$ special placeholder tokens (e.g., ``<SP\_1>'' for the first ID token), visualized as cyan rectangles, to represent the positions of the ground-truth semantic ID. This unique prompt structure enables a critical optimization: with a single prefill pass, the model generates $L+1$ hidden states: one for the context and one for each placeholder.
These hidden states are then fed into the draft head. The draft head consists of two components: a logit head (the yellow diamond), which predicts token probabilities, and a transition module (the gray step), which updates the context representation after each new token is generated. The behavior of this head differs between training and inference: During training, it learns to predict the known ground-truth tokens from the offline data ($\mathcal{L}$). During inference, it performs a beam search, generating the top-$K$ candidate tokens at each step. 
After $L$ iterative calls to the draft head, a pool of candidate items $\hat Y_L$ is generated. Finally, \name performs a fast, model-free verification. Instead of invoking the target LLM, it simply checks the generated sequences against a pre-computed hash set of all valid item IDs $\mathcal{V}$. The top-$K$ valid sequences with the highest cumulative probabilities are returned as the final recommendations. The figure provides an example where two of four candidates are valid (marked with checks).

\subsection{Nimble Drafting}
This section constructs the drafting framework for \name, including the prompting and the autoregressive draft head. 

Existing SD applications typically rely on an external draft model~\cite{zagyva2025speed,xi2025efficiency,lin2025efficient}, which introduces significant maintenance overhead and deployment complexity. To circumvent these issues, we adopt a self-drafting architecture~\cite{xia2024unlocking,cai2024medusa}, which empowers the target LLM itself to generate candidate tokens. 

However, applying general self-drafting architectures like multi-token-prediction (MTP) head~\cite{gloeckle2024betterMTP,cai2024medusa} to GR systems introduces a unique challenge: the highly structured nature of semantic IDs. Unlike free-form text, a semantic ID is a strictly ordered sequence. For an item $v_i$ with three-digit ID $[t^i_1,t^i_2,t^i_3]$, any permutation like $[t^i_3,t^i_2,t^i_1]$ would represent an invalid or entirely different item. 
The token combinations are not arbitrary but are highly constrained, creating strong sequential dependencies. This means that only a sparse subset of all possible sequences is valid. For instance, knowing the prefix $[t^i_1,t^i_2]$ significantly narrows the set of possible values for the final token $t^i_3$ .

Our design for \name directly addresses this structural constraint. By prompting the input with positional placeholder tokens and utilizing an autoregressive draft head, we explicitly provide the model with the positional awareness and sequential modeling capabilities necessary to generate valid, structured IDs.
Specifically, \name first modifies the input prompt for each training instance $\{(q,x),y\}$. Instead of appending the ground-truth semantic ID $y=[t^y_1,t^y_2,t^y_3]$, we append a sequence of $L$ special tokens as the placeholder for item ID: $[\text{<SP\_1>},\dots, \text{<SP\_L>}]$.  %This is visually revealed by the modification from Figure~\ref{fig:intro}(b) to Figure~\ref{fig:frame}.

The purpose of these placeholders is twofold. First, drawing inspiration from~\cite{samragh2025your}, they explicitly encode positional information, prompting the model to leverage its full contexts for each step of the autoregressive generation. Second, they allow the LLM to pre-compute the hidden states for all future token positions in a single forward pass, as visualized in the cyan block of Figure~\ref{fig:frame}. 

We now formally define the autoregressive draft head. Let $\boldsymbol{h}_0,\boldsymbol{h}_1,\dots,\boldsymbol{h}_L$ be the hidden states from the LLM's final layer, corresponding to the prefix ($\boldsymbol{h}_0$) and the $L$ placeholder tokens, respectively. The draft head is defined as follows:
\begin{gather}
\boldsymbol{s}_1 = \boldsymbol{h}_0 \label{eq:init}\\    
\boldsymbol{p}_l=\mathrm{logit\_head}_l(\boldsymbol{h}_{l},\boldsymbol{s}_{l})\label{eq:logit}\\
\boldsymbol{s}_{l+1}=\mathrm{Transition}_l(\boldsymbol{s}_{l},\boldsymbol{e}_l)\label{eq:transition}
\end{gather}
Here, $\boldsymbol{s}_l$ represents the evolving context state, which is initialized by the prefix representation $\boldsymbol{h}_0$. In each step, the $\mathrm{logit\_head}_l$ produces the next-token probability distribution $\boldsymbol{p}_l$. Concurrently, the transition module, $\mathrm{Transition}_l$, updates the context state from $\boldsymbol{s}_l$ to $\boldsymbol{s}_{l+1}$ by incorporating the embedding $\boldsymbol{e}_l$ of the newly selected token $t_l$\footnote{Note that for clarity in this general formulation, we omit the superscripts identifying specific candidates in the beam. This process is detailed further in Section~\ref{sec:trainandinfer}.
}, achieving the autoregressive prediction.

\begin{algorithm}[ht]
	\caption{The \name Pipeline: Training and Inference.}\label{alg:nezha}
	\raggedright
    \textbf{\underline{Training Process}} \\
        {\bf Input}: LLM $\mathcal{M}_p$ with the draft head, training data $\mathcal{D}=\{(q,x),y\}$, length of semantic ID $L$\\
	{\bf Output}: Optimized LLM $\mathcal{M}_p$ with the trained draft head\\
	\begin{algorithmic} [1] 
     \FOR{$(q,x),y \in \mathcal{D}$}
    \STATE Prefilling context and placeholders with LLM: $\boldsymbol{h}_0,\boldsymbol{h}_1,\dots,\boldsymbol{h}_L=\mathcal{M}_p(q,x,<SP\_1>,\dots,<SP\_L>)$
      \STATE Initialization: $l=1, \boldsymbol{s}_1=\boldsymbol{h}_0$
       \FOR{$l\leq L$}
            \STATE Calculating the probabilities $\boldsymbol{p}_l$ of the next token according to Equation~\eqref{eq:logit}
            \STATE Selecting the probability $p^y_l$ and obtaining the embedding $\boldsymbol{e}^y_l$ for the ground-truth token $t^y_l$
            \STATE Updating the context state: $\boldsymbol{s}_{l+1}\gets \mathrm{Transition}_{l}(\boldsymbol{s}_{l},\boldsymbol{e}^y_l)$
        \ENDFOR
        \STATE Calculating the loss $\mathcal{L}$ according to Equation~\eqref{eq:loss}
        \STATE Updating the parameter of $\mathcal{M}_p$ and the draft head ($\mathrm{Transition}_l$ and $\mathrm{logit\_head}_l$)
        \ENDFOR
    \RETURN $\mathcal{M}_p$ with the trained draft head
	\end{algorithmic}
    \textbf{\underline{Inference Process}} \\
    {\bf Input}: LLM $\mathcal{M}_p$ with the trained draft head, query $q$, user sequence $x$, beam size $K$, length of semantic ID $L$, index set for valid semantic IDs $\mathcal{V}$\\
	{\bf Output}: $\hat Y$ (top $K$ items represented by semantic IDs)\\
    	\begin{algorithmic} [1]
         \setcounter{ALC@line}{12}
      \STATE Prefilling context and placeholders with LLM: $\boldsymbol{h}_0,\boldsymbol{h}_1,\dots,\boldsymbol{h}_L=\mathcal{M}_p(q,x,<SP\_1>,\dots,<SP\_L>)$
      \STATE Initialization: $l=1, \boldsymbol{s}_1=\boldsymbol{h}_0,\,\hat Y_0=\{\varnothing\},\mathcal{S}_0=\{\boldsymbol{s}_1\} $
 \FOR{$l\leq L$}
 \STATE $\hat Y_l,\mathcal{S}_l\gets \{\},\{\}$
\FOR{$\hat y \in \hat Y_{l-1}, \boldsymbol{s}_l\in \mathcal{S}_{l-1}$}
     \STATE Calculating the probabilities $\boldsymbol{p}_l$ of the next token according to Equation~\eqref{eq:logit}
     \STATE Selecting $K$ tokens with top probabilities: $\{t^k_l, p^k_l\}_{k\leq K}$
     
     \STATE Obtaining the embedding $\boldsymbol{e}^k_l$ for each beam $t^k_l$, $\boldsymbol{s}_{l+1}\gets \mathrm{Transition}_{l}(\boldsymbol{s}_{l},\boldsymbol{e}^k_l),$
    \STATE $\hat Y_l \gets  \hat Y_l \cup [\hat y, t^k_l], \mathcal{S}_l\gets \mathcal{S}_l\cup \boldsymbol{s}_{l+1}, \forall k$
            \ENDFOR
        \IF{l==L}
        \STATE Calculating the index $\mathcal{V}_L$ for $K\times K$ candidates in $\hat Y_L$ according to Equation~\eqref{eq:index}
        \STATE Retaining valid predictions according to Equation~\eqref{eq:verification}
        \ENDIF 
        \STATE Keeping $K$ candidates in $\hat Y_l$ and corresponding states $\mathcal{S}_l$ with the largest cumulative probability ($\prod_{t^k_l\in\hat y}p_l^k, \hat y \in \hat Y_l$)
        \STATE $l\gets l +1$
        \ENDFOR
\RETURN $\hat Y_L$
	\end{algorithmic}
\end{algorithm}

\subsection{Efficient Verification}
We elaborate on the model-free verification in this section.

Conventional SD verification requires a computationally expensive forward pass of the target model $\mathcal{M}_p$ to evaluate complex grammar and semantics~\cite{cai2024medusa,leviathan2023fast,li2024eagle}. This step can be a primary latency bottleneck in online systems. 

% The highly structured semantic IDs, which are special tokens that irrelevant with explicit grammar and semantics, allows for a far more efficient verification without any model invocation. We propose a model-free verification approach that leverages the strict combinatorial rules of valid item IDs. Instead of a forward pass, we perform a simple, highly-efficient lookup to check if a generated sequence corresponds to an actual item in our database.

The highly structured nature of semantic IDs, which are defined by strict combinatorial rules rather than linguistic grammar, enables a far more efficient verification process. This key property allows us to verify a candidate sequence's validity by performing a near-instantaneous lookup against a precomputed set of all valid item IDs, rather than executing a costly forward pass. The power of this approach stems from the extreme sparsity of the valid ID space. As supported by both public datasets~\cite{wang2024learnable} and our own industrial practice, the ratio of valid IDs is exceptionally low (around 0.1\% in our system). This means our simple verification step can correctly filter out over 99.9\% of all drafted candidates. 
%For instance, given two candidate sequences with similar probabilities, such as $(243, 129, 3)$ and $(0, 346, 122)$, our method can instantly discard the latter if it does not map to any valid item, thereby promoting the correct candidate. 

To implement this model-free verification with minimal computational overhead, we first encode each multi-token semantic ID into a single, unique integer~\cite{zheng2025enhancing}:
\begin{gather}
    V_i = \mathcal{P}(v_i)=\sum_{l=1}^L (t_l^i\times\prod_{j=1}^{l-1}T_j)\label{eq:index}
\end{gather}
where this equation maps the semantic ID $[t_1^i,\dots,t_L^i]$ to a unique integer index $V_i$ using a mixed-radix conversion. $T_j$ is the vocabulary size for the $j$-th token position. The product term $\prod_{j=1}^{l-1}T_j$ acts as a positional multiplier. For the first token ($l=1$), this product is empty and set to 1 by convention to correctly initialize the formula. 

Equation~\eqref{eq:index} provides a bijective mapping from this multi-dimensional token space to a one-dimensional integer space, assigning a unique index to every possible ID within the range $[0,\prod_{l=1}^{L}T_l-1]$. For example, consider the three-token ID $(243, 129, 3)$ from Figure~\ref{fig:frame}, and assume the vocabulary size for each token position is $512$. The unique integer index is calculated as: $243\times1+129\times 512+3\times (512 \times 512) = 852723$. 

Let $\mathcal{V}$ be the set of all valid semantic IDs, implemented as a hash set for efficient, constant-time lookups. A batch of predicted sequences from the LLM can then be verified by computing the intersection with this set of valid IDs:
\begin{gather}
    \hat Y_L = \hat Y_L[\mathcal{V}\cap\mathcal{V}_L] \label{eq:verification}
\end{gather}
where $\mathcal{V}_L$ is the set of integer indices corresponding to the candidate items in the final step $\hat Y_L$. This set is then filtered in-place according to the verification process detailed in Equation~\eqref{eq:verification}, retaining only the indices of valid items, i.e., with indices included in $\mathcal{V}\cap\mathcal{V}_L$. The visualized example in Figure~\ref{fig:frame} illustrates this process, where $(0,346,122), (12,1,33)$ is included by the updated $\hat Y_L$ while excluding other candidates.

The impact of this model-free verification is substantial. On our production data, it boosts the ratio of valid drafted candidates from just 43\% to over 93\%. This directly translates into a 12 percentage-point uplift in key offline evaluation metrics. A detailed component-wise analysis using public datasets to confirm this contribution is provided in Section~\ref{sec:ab}.

\subsection{Training and Inference}~\label{sec:trainandinfer}
This section details the pipeline of \name, summarized in Algorithm~\ref{alg:nezha}, including the training and inference processes. While both processes share the architecture, their objectives differ, leading to distinct approaches for token selection and state transition.

During training, the model learns to predict the ground-truth semantic ID using a teacher-forcing approach. For a given training instance with label $y=[t^y_1,\dots,t^y_L]$: LLM first initializes the representations with a single call(line 2). At each step $l$, the model uses Equation~\eqref{eq:logit} to compute the probability distribution $\boldsymbol{p}_l$ (line 5). For the subsequent transition step (Equation~\eqref{eq:transition}), the context $\boldsymbol{s}_l$ is updated using the embedding of the ground-truth token, i.e., $\boldsymbol{e}_l=\boldsymbol{e}^y_l$ (line 6-7). This ensures the model learns the correct sequential path, regardless of its own predictions at step $l$. Finally, the probability for the ground-truth token, $p^y_l$, is used to compute the cross-entropy loss, which would be minimized over $\mathcal{D}$ (line 9-10):
\begin{gather}
    \mathcal{L}=\sum_{y\in\mathcal{D}}\sum_{l=1}^L \log p_l^y \label{eq:loss}
\end{gather}
The model parameters are optimized by minimizing the loss $\mathcal{L}$ using a gradient-based optimizer (e.g., Adam~\cite{kingma2014adam}). This training procedure yields the final model as in line 12.

In contrast, the goal of inference is to find the most probable semantic IDs, which we achieve using beam search. 
The process begins by prefilling the specialized prompt (line 13). Then, we initialize the step index $l$ and the context state $\boldsymbol{s}_{1}$, and set up the prediction set $\hat Y_0$ and the state set $\mathcal{S}_0$ to track the candidate beams and their respective states (line 14). 
For each step, we iterate over each candidate beam (line 17), compute the probability distribution $\boldsymbol{p}_l$, and select the top $K$ tokens $t^k_{l}$ with the highest probabilities $p^k_{l}$ (lines 18-19).  
The context state for each beam is then updated independently using the embedding of its selected token $\boldsymbol{e}^k_l$ as in line 20. New beams are finally included to $\hat Y$ and $\mathcal{S}_l$ (line 21) for further pruning (line 27), where the top $K$ overall sequences are retained at each step based on their cumulative probabilities. % This process continues autoregressively for the remaining steps ($l\geq2$), with each of the $K$ beams generating its own predictions and updating its own state. 
Notably, the final decoding step ($l=L$) includes a crucial verification stage to filter out hallucinated (i.e., invalid) semantic IDs, rather than simply selecting the top candidates by probability (lines 23-26).

% Each of the $K$ beams from the previous step generates its top $K$ next-token predictions, resulting in a pool of $K\times K$ candidate sequences. These candidates are then passed through our efficient, model-free verification to identify the subset of valid semantic IDs.
% From this valid subset, the top $K$ sequences with the highest cumulative probabilities are finally returned as the final output (line 27). 

% We apply the verification step only at the end because we empirically observed that hallucination rates are negligible in early steps but increase sharply at the final token. However, given the near-zero cost of our lookup-based verification, it could be applied at every step to ensure validity across the inference without a significant efficiency penalty.

\begin{table}[t]
\centering
\caption{The statistics of the preprocessed datasets}
\begin{tabular}{ccccc}
\toprule[1pt]
\textbf{Dataset} & \textbf{\# Users} & \textbf{\# Items} & \textbf{\# Intersections} & \textbf{Avg.length} \\ 
\midrule
Yelp & 15,719 & 11,383 & 192,214 & 12.22 \\
Beauty & 52,204 & 57,288 & 394,908 & 7.56 \\
Games & 64,071 & 33,614 &568,314  & 8.87 \\ 
\bottomrule[1pt]
\end{tabular}
\label{tab:exp_dataset}
% \vspace{-4mm}
\end{table}

\begin{table*}[ht]
\tabcolsep=0.1cm 
\centering
\caption{Overall performance of \name. The boldface refers to beating all baselines. ``\textbf{{\Large *}}'' indicates the statistically significant improvements (i.e., one-sided t-test with $p<0.05$). For performance metrics, the higher is better. For ``LT'', the lower is better.}
\label{tab:ovall}
\resizebox{\textwidth}{!}{%
\begin{tabular}{@{}c|c|ccccc|ccccc|ccccc@{}}
\toprule
 \multirow{2}{*}{Model} & \multirow{2}{*}{Finetuning}                 & \multicolumn{5}{c|}{Beauty} & \multicolumn{5}{c|}{Games}& \multicolumn{5}{c}{Yelp} \\ 
&  & H@5  & H@10  & N@5  & N@10  & LT& H@5  & H@10 & N@5 & N@10& LT & H@5  & H@10 & N@5 & N@10& LT \\ \midrule
\multirow{3}{*}{Llama}                 
& Beam Search&0.0392&0.0568&0.0254&0.0311&74.21&0.0305&0.0406&0.0224&0.0256&78.95&0.0176&0.0262&0.0110&0.0138&75.99\\
& MTP&0.0378&0.0555&0.0239&0.0297&6.74&0.0291&0.0400&0.0212&0.0248&7.33&0.0140&0.0240&0.0091&0.0114&7.18 \\
% & Medusa&\\
% & AtSpeed &\\
&\name &0.0390&0.0562&0.0253&0.0304&7.00&\textbf{0.0313}&\textbf{0.0410}&\textbf{0.0231}&\textbf{0.0259}&7.52&\textbf{0.0194*}&\textbf{0.0332*}&\textbf{0.0125*}&\textbf{0.0168*}&7.41  
\\\midrule
\multirow{3}{*}{{Qwen}}       
& Beam Search& 0.0446&0.0643&0.0306&0.0370&47.70&0.0226&0.0305&0.0153&0.0179&50.06&0.0235&0.0377&0.0147&0.0192&48.31  \\
& MTP&0.0422&0.0617&0.0287&0.0352&4.18&0.0204&0.0283&0.0136&0.0170&4.61&0.0229&0.0346&0.0114&0.0177&4.24 \\
% & Medusa&\\
% & AtSpeed&\\
&\name &\textbf{0.0451}&\textbf{0.0649}&\textbf{0.0308}&\textbf{0.0377}&4.41&\textbf{0.0239*}&\textbf{0.0314}&\textbf{0.0165*}&\textbf{0.0193*}&4.80&0.0235&0.0363&0.0143&0.0185&4.46             
\\\bottomrule
\end{tabular}%
% \vspace{-5mm}
}
\end{table*}

\section{Experiment}
To demonstrate the generalizability and robustness of \name, this section presents an evaluation on public benchmark datasets.
\subsection{Settings}
\textbf{Datasets.}
We evaluate \name on three public datasets widely used as benchmarks for GR: Yelp, Amazon Beauty, and Amazon Games~\cite{wang2024learnable,zheng2024adapting}. These datasets are particularly suitable as they contain rich contextual information, such as textual item descriptions.
Following standard practice in prior work, we employ a leave-one-out strategy for data splitting. For each user, the last interaction is reserved for the test, the penultimate interaction is used for validation, and remaining interactions constitute the training set.

\noindent\textbf{Baselines and Backbones.} We construct the generative recommender with two backbones, Llama-1B~\cite{dubey2024llama} and QWen3-0.8B~\cite{yang2025qwen3}, which have a similar size to the LLM we used in our production environments. We primarily compare with two baselines on public datasets: vanilla LLM decoding with beam search and MTP~\cite{gloeckle2024betterMTP}. Other baselines, such as Medusa~\cite{cai2024medusa,zagyva2025speed} and AtSpeed~\cite{lin2025efficient}, are omitted due to their excessive latency and complexity for practical use. 

% Medusa~\cite{cai2024medusa}, and AtSpeed~\cite{lin2025efficient}. 
% We compare \name with three categories of LLM fine-tuning baselines: (a) vanilla \textbf{SFT}; (b) on-policy RLFT: \textbf{PPO}~\cite{schulman2017proximalPPO}, \textbf{GRPO}~\cite{shao2024deepseekmathGRPO}; (c) off-policy RLFT: \textbf{DPO}~\cite{rafailov2023directDPO}, and DPO-based algorithms modified for language-based recommendations, including \textbf{S-DPO}~\cite{chensoftmax}, \textbf{SPRec}~\cite{gao2025sprec}, \textbf{IPA}~\cite{deng2025onerec}. Notably, the specially designed auxiliary loss in addition to SFT is regarded as a part of the GR backbones. And we install \name to two backbones: \textbf{TIGER}~\cite{rajput2023recommender} and \textbf{LETTER}~\cite{wang2024learnable}.

\noindent\textbf{Evaluation Metrics.}
We evaluate all methods using Hit Rate (HR) and Normalized Discounted Cumulative Gain (NDCG) at cutoffs of 5 and 10, yielding four accuracy metrics: \textit{H@5}, \textit{H@10}, \textit{N@5}, and \textit{N@10}. For efficiency, we also report the average generation latency (LT) in milliseconds (ms), which is the sum of the ``Prefill'' and ``Decode'' costs shown in Table~\ref{tab:latdec}. 

\noindent\textbf{Implementation Details.}
The evaluation on public datasets is conducted using the same GPU devices to ensure consistency. To achieve robust results, all experiments are averaged over three runs with distinct random seeds (42, 43, 44). For item tokenization, we set  $L=3$ with $T_l=512$ for each layer, meaning the codebook size is 512 across all layers. Item tokenization is performed using RQ-VAE~\cite{lee2022autoregressive}. These settings are consistent with our production environment, with the exception of $T_l$, which varies due to the different volumes of items.
We convert item attributes into textual instructions and obtain item representations through the public API, such as text-embedding-ada-002\footnote{\url{https://platform.openai.com/docs/guides/embeddings}}. The backbone LLM is determined by the specific setting, either Llama or Qwen, and is optimized using Transformer Trainers, with the appropriate hyperparameters chosen for each test.
In Equation~\eqref{eq:logit}, the $\mathrm{logit\_head}_l$ represents a linear transformation layer shaped $[d_{hid}, T_l]$, which is used to convert the hidden state into logits; specifically, this is sized at $1024 \times 512$ for each layer in our case. Additionally, we employ an RNN to update the context state, i.e., $\mathrm{Transition}_l$ in Equation~\eqref{eq:transition}. For inference, the beam size $N$ is set as $10$.

\subsection{Overall Performance}
We test \name with two LLM backbones and compare it with the original beam search and efficient decoding baseline to evaluate its effectiveness. The results are listed in Table~\ref{tab:ovall}, which consistently \name presents the satisfactory performance with outstanding efficiency. We can observe that:
\begin{itemize}[leftmargin=*]
    \item \textbf{Comparison with Beam Search. } The performance of \name is comparable to that of beam search but with a significant increase in speed, achieving approximately a 10-fold improvement. This demonstrates its potential to effectively replace beam search for online GR serving. In certain instances, such as using Llama on Yelp, \name even produces enhanced results.
    \item \textbf{Comparison with MTP. } Although MTP is more efficient than \name in terms of lower latency (LT), it fails to deliver satisfactory performance due to its disregard for the characteristics of GR, specifically the structured semantic IDs. The parallel decoding approach used by MTP, which relies on a shared last hidden state, cannot leverage the sequential dependency inherent in the semantic ID, resulting in a failure to predict different tokens with discriminative representations.
\end{itemize}
% \usepackage{booktabs}
\begin{table}[]

\caption{Ablation study on Yelp with Llama.}
\label{tab:abl}
\begin{tabular}{@{}c|ccccc@{}}
\toprule
        & H@5    & H@10   & N@5    & N@10   & LT     \\
        \midrule
NEZHA   & 0.0194 & 0.0332 & 0.0125 & 0.0168 & 7.41\\
NEZHA-1 & 0.0044 & 0.0083 & 0.0024 & 0.0037 & 7.03 \\
NEZHA-2 & 0.0188 & 0.0309 & 0.0120 & 0.0159 & 7.41 \\
NEZHA-3 & 0.0173 & 0.0322 & 0.0110 & 0.0158 & 7.18 \\
NEZHA-4 & 0.0194 & 0.0328 & 0.0124 & 0.0167 & 7.16 \\\bottomrule
\end{tabular}
\end{table}

\subsection{Ablation Study}\label{sec:ab}
To validate the contribution of each key component in \name, we conduct an ablation study with the following four variants:
\begin{itemize}[leftmargin=*]
   \item  \textbf{\name-1:} This variant removes the context state $\boldsymbol{s}_l$ from the logit calculation in Equation~\eqref{eq:logit}, which also implies the removal of the transition head $\mathrm{Transition}_l$ (Equation~\eqref{eq:transition}), making the prediction of each token within a semantic ID independent of the previously generated tokens. 
   \item \textbf{\name-2:} This variant removes the placeholder hidden state $\boldsymbol{h}_l$ from Equation~\eqref{eq:logit}. This allows us to measure the importance of providing positional information for effective self-drafting.
   \item \textbf{\name-3: }This variant replace the RNN with simple addition for $\mathrm{Transition}_l$ to investigate the impact of state transition modeling. 
   \item \textbf{\name-4:} This variant disables the model-free verification step entirely. It is designed to demonstrate the crucial trade-off between efficiency and accuracy. % showing that verification provides a significant performance lift with a negligible increase in latency.
\end{itemize}
We evaluate these variants using the Llama backbone on Yelp, with the results summarized in Table~\ref{tab:abl}. We find the following insights:

\begin{itemize}[leftmargin=*]
    \item \name-1 suffers from performance collapse due to the absence of context and sequential dependency. This underlines the inadequacy of placeholder representations for context understanding, highlighting the necessity of maintaining a context state.
    \item \name-2 performs slightly worse than the original \name, which can be attributed to the loss of placeholders. When combined with the performance of \name-1, it is evident that placeholders can enhance performance by providing positional information for semantic IDs, although they are not indispensable.
    \item The simplification of the transition module negatively impacts the performance of \name-3, indicating the importance of accurately modeling state transitions to leverage the sequential dependency patterns of GR.
    \item A comparison of \name-4 with \name shows that the slight increase in latency for verification (0.25 ms) can lead to improved performance (especially on H@10), emphasizing the effectiveness of our verification process.
\end{itemize}
\begin{figure}[t]
\centering
% \vspace{-1mm}
\includegraphics[width=0.95\linewidth]{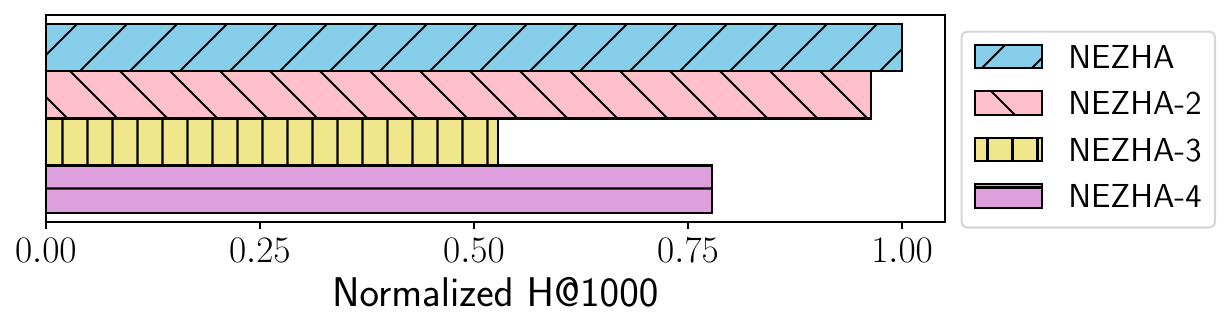}
\caption{Ablation study on production data. We present the normalized performance for confidentiality.}
\label{fig:abl}
\vspace{-3mm}
\end{figure}
Additionally, we present the ablation results on offline production data in Figure~\ref{fig:abl} to further validate our conclusions. Notably, \name consistently outperforms all variants. However, trends differ across specific variants compared with public data, which may stem from variations in data distribution and volume. For instance, the volume of items significantly increases in production data, resulting in greater performance degradation for \name-3 and \name-4. This degradation can be driven by weakening the sequential dependency and by introducing too many invalid predictions, respectively. Meanwhile, \name-1 and \name-2 show the consistent pattern as on public datasets: \name-1 is omitted from the figure due to its limited performance, while \name-2 is only slightly worse than the original version.

% \subsection{}

\section{Real-world Deployment}
\begin{figure}[t]
\centering
% \vspace{-1mm}
\includegraphics[width=0.9\linewidth]{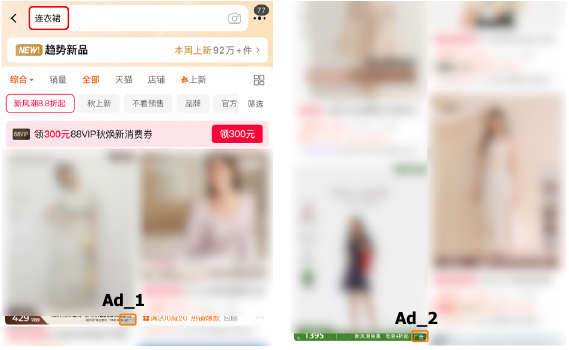}
\caption{Illustration of the method's deployment in Taobao Search. The example shows results for the query ``Dress'' (top left, in the red box). Advertisements are identified by an orange label in the bottom-right corner of each product. Sensitive information has been obscured for confidentiality.}
\label{fig:dpl}
% \vspace{-3mm}
\end{figure}

The proposed method, \name, is deployed in the candidate generation (recall) stage of the \textbf{Taobao Search Advertising} platform. The operational context is illustrated in Figure~\ref{fig:dpl}, which displays search results for the query ``Dress''. Advertisements displayed on the first page (``Ad\_1'') are subject to a stringent latency constraint of 30ms. Previously, our GR system's inference time of over 1000ms precluded its use for these prime positions, limiting its deployment to subsequent pages (``Ad\_2'').

With the introduction of \name, we overcome this critical limitation. As detailed in Table~\ref{tab:latdec}, \name provides a \textbf{2.6$\times$} algorithm-level speedup. This acceleration, coupled with system-level savings on queuing time, reduces the GR system's total latency from over 1000ms to below the 30ms threshold. Consequently, we can now deploy the GR system across all advertising slots, including the highly valuable first-page placements.

The efficacy of this deployment is confirmed through extensive experiments. Offline evaluations show an absolute increase in the hit rate on clicked items by 0.58\% and 0.61\% for the top 500 and 1000 results, respectively. The positive result enables the deployment of the GR system in the prime position as of October 2025, where 7-day reverse online A/B testing on 10\% traffic registered a \textbf{1.2\%} revenue increase (\textbf{billion-level} improvement).

%% 单栏表
% %%% Statistics of Datasets %%%
% \begin{table}[]
% \centering
% \caption{The Statistics of Datasets}
% \begin{tabular}{ccccc}
% \toprule[1.5pt]
% Dataset & \# Users & \# Items & Sparsity & Avg.length \\ 
% \midrule
% \midrule
% Yelp & 30,431 & 20,033 & 99.95\% & 10.40 \\
% Beauty & 50,498 & 57,289 & 99.99\% & 7.76 \\
% Steam & 332,925 & 13,047 & 99.91\% & 11.06 \\ 
% \bottomrule[1.5pt]
% \end{tabular}
% \label{tab:dataset}
% \vspace{-4mm}
% \end{table}
% %%% Statistics of Datasets %%%

\section{Related Works}
Our work is situated at the intersection of two rapidly evolving research areas: generative recommendations, which leverage the power of LLMs for recommendation tasks, and speculative decoding, a key technique for accelerating LLM inference.

\noindent\textbf{Generative Recommendations. }  
% Traditional recommender systems, dominated by collaborative filtering and factorization models~\cite{}, have recently been challenged by the emergence of LLMs. Generative Recommendation (GR) reframes the recommendation task from a discriminative problem (predicting scores or ranking items) to a generative one (generating a set of next items to recommend)~\cite{rajput2023recommender}, which has unlocked several advantages, including the ability to handle cold-start items naturally, incorporate rich textual side information, and enhance the diversity.
Significant research has been devoted to advancing each stage of the GR pipeline, i.e., item tokenization, LLM training, and LLM serving.
For item tokenization, beyond using multi-modal content, recent work has started to incorporate collaborative information to create more effective semantic IDs~\cite{wang2024learnable,zhu2024cost}. The design of these IDs has also been extended to more complex scenarios, such as multi-domain~\cite{zheng2025universal} and multi-behavior~\cite{liu2024multi} recommendations, while other approaches seek to unify the training of the tokenizer and the LLM in an end-to-end framework~\cite{liu2024end}.
For training, researchers have explored pre-training paradigms to align item semantics with the LLM's linguistic space~\cite{zheng2024adapting}, as well as post-training techniques like Direct Preference Optimization (DPO)~\cite{rafailov2023directDPO} to fine-tune model outputs toward specific business objectives~\cite{deng2025onerec,wei2025oneloc}.
To facilitate serving, researchers adapted speculative decoding for top-k recommendation~\cite{lin2025efficient}, enabled out-of-vocabulary inference with SpecGR~\cite{ding2024inductive}, and reformulated the system with long semantic IDs and parallel decoding in RPG~\cite{hou2025generating}.

Despite the impressive performance, the practical deployment of GR models is severely hampered by high inference latency. The autoregressive generation requires multiple forward passes through LLMs, a process that is often too slow for industrial systems demanding real-time responses, motivating the proposal of \name.

\noindent\textbf{Speculative Decoding. } 
To address the high latency associated with autoregressive generation, Speculative Decoding (SD) has emerged as a leading technique for accelerating inference~\cite{leviathan2023fast,xia2024unlocking,zhang2025surveyparallel}. The standard SD framework utilizes a small, fast ``draft model'' to generate a sequence of candidate tokens, which is then verified in a single parallel forward pass by the larger, original ``target model''.

Subsequent research has explored various strategies to enhance both stages of this process. For instance, Medusa~\cite{cai2024medusa} introduces multiple lightweight decoding ``heads'' attached to the target model to predict future tokens in parallel~\cite{gloeckle2024betterMTP}, and it proposes tree-attention verification to validate drafts with just one call to the large model. Some researchers have replaced the parallel decoding heads with autoregressive heads, shifting from parallel multi-token drafting to token-by-token generation~\cite{li2024eagle,cheng2024recurrent}. Additionally, researchers also suggest using retrieval models as the draft model instead of relying solely on language models~\cite{he2024rest}. These innovations have led to successful deployments in various industrial applications, including travel services~\cite{zagyva2025speed} and knowledge-based recommendations~\cite{xi2025efficiency}.

However, current practices still rely on separate draft models and require verification by the large target model. Our approach pioneers the exploration of self-drafting in large-scale production environments and introduces an instant and effective model-free verification specialized for GR.

\section{Conclusion}
% In this paper, we address the critical latency challenges of deploying Large Language Models (LLMs) for industrial-scale Generative Recommendation (GR). Existing acceleration techniques like KV-Caching, might be insufficient for GR. While existing practical Speculative Decoding (SD) solutions, often fall short due to their reliance on auxiliary draft models and costly verification steps, rendering them impractical for production environments with stringent real-time requirements. To bridge this critical gap, we propose \name, a novel and highly efficient framework for generative recommendation. By employing a representation system of placeholders and context states, we enable the main LLM to produce the hidden representation in parallel within a single, efficient prefill phase, subsequently generate multiple candidates with the autoregressive head. This self-dratfing manner entirely eliminates the need for an independent, smaller draft model. Additionally, we leverage the highly structured nature of semantic IDs in GR to replace the costly LLM-based validation with a pre-computed lookup structure (e.g., a hash set). This renders the verification step near-instantaneous and ensures all final outputs are valid by construction.
% Extensive experiments on several public benchmark datasets demonstrate the superior performance of \name. More importantly, it significantly reduces generation latency to levels required by industrial applications, benefiting the billion-level business.
Industrial-scale Generative Recommendation (GR) using Large Language Models (LLMs) is severely constrained by high inference latency. Existing acceleration methods, including Speculative Decoding (SD), are often impractical due to their reliance on auxiliary draft models and expensive verification steps. To overcome these limitations, we propose \name, a highly efficient decoding framework for GR. \name introduces two key innovations: (1) self-drafting, which allows the main LLM to generate multiple candidates in a single pass, eliminating the need for a separate draft model; and (2) model-free verification, which uses a near-instantaneous hash lookup to validate candidates, bypassing costly model-based checks. Experiments show that \name achieves zero-sacrifice accuracy while reducing latency to levels suitable for large-scale deployment.
% \begin{acks}
% \end{acks}

\begin{acks}
This research was partially supported by National Natural Science Foundation of China (No.62502404), Hong Kong Research Grants Council (Research Impact Fund No.R1015-23, Collaborative Research Fund No.C1043-24GF, General Research Fund No.11218325), Institute of Digital Medicine of City University of Hong Kong (No.9229503), and Alibaba (CCF-Alimama Tech Kangaroo Fund No. 2024002).
\end{acks}


\clearpage
\bibliographystyle{ACM-Reference-Format}
\bibliography{reference}

@inproceedings{xi2024KAR,
  title={Towards open-world recommendation with knowledge augmentation from large language models},
  author={Xi, Yunjia and Liu, Weiwen and Lin, Jianghao and Cai, Xiaoling and Zhu, Hong and Zhu, Jieming and Chen, Bo and Tang, Ruiming and Zhang, Weinan and Yu, Yong},
  booktitle={Proceedings of the 18th ACM Conference on Recommender Systems},
  pages={12--22},
  year={2024}
}

@inproceedings{ren2024representation,
  title={Representation learning with large language models for recommendation},
  author={Ren, Xubin and Wei, Wei and Xia, Lianghao and Su, Lixin and Cheng, Suqi and Wang, Junfeng and Yin, Dawei and Huang, Chao},
  booktitle={Proceedings of the ACM on Web Conference 2024},
  pages={3464--3475},
  year={2024}
}

@article{bao2025bi,
  title={A bi-step grounding paradigm for large language models in recommendation systems},
  author={Bao, Keqin and Zhang, Jizhi and Wang, Wenjie and Zhang, Yang and Yang, Zhengyi and Luo, Yanchen and Chen, Chong and Feng, Fuli and Tian, Qi},
  journal={ACM Transactions on Recommender Systems},
  volume={3},
  number={4},
  pages={1--27},
  year={2025},
  publisher={ACM New York, NY}
}

@article{rajput2023recommender,
  title={Recommender systems with generative retrieval},
  author={Rajput, Shashank and Mehta, Nikhil and Singh, Anima and Hulikal Keshavan, Raghunandan and Vu, Trung and Heldt, Lukasz and Hong, Lichan and Tay, Yi and Tran, Vinh and Samost, Jonah and others},
  journal={Advances in Neural Information Processing Systems},
  volume={36},
  pages={10299--10315},
  year={2023}
}

@inproceedings{zheng2024adapting,
  title={Adapting large language models by integrating collaborative semantics for recommendation},
  author={Zheng, Bowen and Hou, Yupeng and Lu, Hongyu and Chen, Yu and Zhao, Wayne Xin and Chen, Ming and Wen, Ji-Rong},
  booktitle={2024 IEEE 40th International Conference on Data Engineering (ICDE)},
  pages={1435--1448},
  year={2024},
  organization={IEEE}
}

@article{zhou2025onerec,
  title={OneRec-V2 Technical Report},
  author={Zhou, Guorui and Hu, Hengrui and Cheng, Hongtao and Wang, Huanjie and Deng, Jiaxin and Zhang, Jinghao and Cai, Kuo and Ren, Lejian and Ren, Lu and Yu, Liao and others},
  journal={arXiv preprint arXiv:2508.20900},
  year={2025}
}

@article{zheng2025ega,
  title={EGA: A Unified End-to-End Generative Framework for Industrial Advertising Systems},
  author={Zheng, Zuowu and Wang, Ze and Yang, Fan and Fan, Jiangke and Zhang, Teng and Wang, Xingxing},
  journal={arXiv preprint arXiv:2505.17549},
  year={2025}
}

@article{guo2025onesug,
  title={OneSug: The Unified End-to-End Generative Framework for E-commerce Query Suggestion},
  author={Guo, Xian and Chen, Ben and Wang, Siyuan and Yang, Ying and Lei, Chenyi and Ding, Yuqing and Li, Han},
  journal={arXiv preprint arXiv:2506.06913},
  year={2025}
}

@article{wei2025oneloc,
  title={OneLoc: Geo-Aware Generative Recommender Systems for Local Life Service},
  author={Wei, Zhipeng and Cai, Kuo and She, Junda and Chen, Jie and Chen, Minghao and Zeng, Yang and Luo, Qiang and Zeng, Wencong and Tang, Ruiming and Gai, Kun and others},
  journal={arXiv preprint arXiv:2508.14646},
  year={2025}
}

@article{deng2025onerec,
  title={OneRec: Unifying Retrieve and Rank with Generative Recommender and Iterative Preference Alignment},
  author={Deng, Jiaxin and Wang, Shiyao and Cai, Kuo and Ren, Lejian and Hu, Qigen and Ding, Weifeng and Luo, Qiang and Zhou, Guorui},
  journal={arXiv preprint arXiv:2502.18965},
  year={2025}
}

@article{chen2025onesearch,
  title={OneSearch: A Preliminary Exploration of the Unified End-to-End Generative Framework for E-commerce Search},
  author={Chen, Ben and Guo, Xian and Wang, Siyuan and Liang, Zihan and Lv, Yue and Ma, Yufei and Xiao, Xinlong and Xue, Bowen and Zhang, Xuxin and Yang, Ying and others},
  journal={arXiv preprint arXiv:2509.03236},
  year={2025}
}

@article{yang2025sparse,
  title={Sparse meets dense: Unified generative recommendations with cascaded sparse-dense representations},
  author={Yang, Yuhao and Ji, Zhi and Li, Zhaopeng and Li, Yi and Mo, Zhonglin and Ding, Yue and Chen, Kai and Zhang, Zijian and Li, Jie and Li, Shuanglong and others},
  journal={arXiv preprint arXiv:2503.02453},
  year={2025}
}

@inproceedings{hua2023index,
  title={How to index item ids for recommendation foundation models},
  author={Hua, Wenyue and Xu, Shuyuan and Ge, Yingqiang and Zhang, Yongfeng},
  booktitle={Proceedings of the Annual International ACM SIGIR Conference on Research and Development in Information Retrieval in the Asia Pacific Region},
  pages={195--204},
  year={2023}
}

@inproceedings{wang2024learnable,
  title={Learnable item tokenization for generative recommendation},
  author={Wang, Wenjie and Bao, Honghui and Lin, Xinyu and Zhang, Jizhi and Li, Yongqi and Feng, Fuli and Ng, See-Kiong and Chua, Tat-Seng},
  booktitle={Proceedings of the 33rd ACM International Conference on Information and Knowledge Management},
  pages={2400--2409},
  year={2024}
}

@article{zheng2025universal,
  title={Universal Item Tokenization for Transferable Generative Recommendation},
  author={Zheng, Bowen and Lu, Hongyu and Chen, Yu and Zhao, Wayne Xin and Wen, Ji-Rong},
  journal={arXiv preprint arXiv:2504.04405},
  year={2025}
}

@inproceedings{xia2024unlocking,
  title={Unlocking Efficiency in Large Language Model Inference: A Comprehensive Survey of Speculative Decoding},
  author={Xia, Heming and Yang, Zhe and Dong, Qingxiu and Wang, Peiyi and Li, Yongqi and Ge, Tao and Liu, Tianyu and Li, Wenjie and Sui, Zhifang},
  booktitle={ACL (Findings)},
  year={2024}
}

@article{zhang2025surveyparallel,
  title={A survey on parallel text generation: From parallel decoding to diffusion language models},
  author={Zhang, Lingzhe and Fang, Liancheng and Duan, Chiming and He, Minghua and Pan, Leyi and Xiao, Pei and Huang, Shiyu and Zhai, Yunpeng and Hu, Xuming and Yu, Philip S and others},
  journal={arXiv preprint arXiv:2508.08712},
  year={2025}
}

@article{wanefficientllm,
  title={Efficient Large Language Models: A Survey},
  author={Wan, Zhongwei and Wang, Xin and Liu, Che and Alam, Samiul and Zheng, Yu and Liu, Jiachen and Qu, Zhongnan and Yan, Shen and Zhu, Yi and Zhang, Quanlu and others},
  journal={Transactions on Machine Learning Research}
}

@article{zhang2024beyond,
  title={Beyond the speculative game: A survey of speculative execution in large language models},
  author={Zhang, Chen and Liu, Zhuorui and Song, Dawei},
  journal={arXiv preprint arXiv:2404.14897},
  year={2024}
}

@article{xi2024decoding,
  title={A decoding acceleration framework for industrial deployable LLM-based recommender systems},
  author={Xi, Yunjia and Wang, Hangyu and Chen, Bo and Lin, Jianghao and Zhu, Menghui and Liu, Weiwen and Tang, Ruiming and Zhang, Weinan and Yu, Yong},
  journal={arXiv e-prints},
  pages={arXiv--2408},
  year={2024}
}

@inproceedings{leviathan2023fast,
  title={Fast inference from transformers via speculative decoding},
  author={Leviathan, Yaniv and Kalman, Matan and Matias, Yossi},
  booktitle={International Conference on Machine Learning},
  pages={19274--19286},
  year={2023},
  organization={PMLR}
}

@inproceedings{cai2024medusa,
  title={Medusa: Simple LLM Inference Acceleration Framework with Multiple Decoding Heads},
  author={Cai, Tianle and Li, Yuhong and Geng, Zhengyang and Peng, Hongwu and Lee, Jason D and Chen, Deming and Dao, Tri},
  booktitle={International Conference on Machine Learning},
  pages={5209--5235},
  year={2024},
  organization={PMLR}
}

@article{li2024eagle,
  title={Eagle: Speculative sampling requires rethinking feature uncertainty},
  author={Li, Yuhui and Wei, Fangyun and Zhang, Chao and Zhang, Hongyang},
  journal={arXiv preprint arXiv:2401.15077},
  year={2024}
}

@inproceedings{gloeckle2024betterMTP,
  title={Better \& faster large language models via multi-token prediction},
  author={Gloeckle, Fabian and Idrissi, Badr Youbi and Rozi{\`e}re, Baptiste and Lopez-Paz, David and Synnaeve, Gabriel},
  booktitle={Proceedings of the 41st International Conference on Machine Learning},
  pages={15706--15734},
  year={2024}
}

@article{cheng2024recurrent,
  title={Recurrent drafter for fast speculative decoding in large language models},
  author={Cheng, Yunfei and Zhang, Aonan and Zhang, Xuanyu and Wang, Chong and Wang, Yi},
  journal={arXiv preprint arXiv:2403.09919},
  year={2024}
}

@article{samragh2025your,
  title={Your LLM Knows the Future: Uncovering Its Multi-Token Prediction Potential},
  author={Samragh, Mohammad and Kundu, Arnav and Harrison, David and Nishu, Kumari and Naik, Devang and Cho, Minsik and Farajtabar, Mehrdad},
  journal={arXiv preprint arXiv:2507.11851},
  year={2025}
}

@inproceedings{
lin2025efficient,
title={Efficient Inference for Large Language Model-based Generative Recommendation},
author={Xinyu Lin and Chaoqun Yang and Wenjie Wang and Yongqi Li and Cunxiao Du and Fuli Feng and See-Kiong Ng and Tat-Seng Chua},
booktitle={The Thirteenth International Conference on Learning Representations},
year={2025}
}

@article{ding2024inductive,
  title={Inductive Generative Recommendation via Retrieval-based Speculation},
  author={Ding, Yijie and Hou, Yupeng and Li, Jiacheng and McAuley, Julian},
  journal={arXiv preprint arXiv:2410.02939},
  year={2024}
}

@article{zagyva2025speed,
  title={Speed without sacrifice: Fine-tuning language models with Medusa and knowledge distillation in travel applications},
  author={Zagyva, Daniel and Stergiadis, Emmanouil and van der Maas, Laurens and Dokic, Aleksandra and Fainman, Eran and Gusev, Ilya and Beladev, Moran},
  year={2025}
}

@inproceedings{xi2025efficiency,
  title={Efficiency unleashed: Inference acceleration for LLM-based recommender systems with speculative decoding},
  author={Xi, Yunjia and Wang, Hangyu and Chen, Bo and Lin, Jianghao and Zhu, Menghui and Liu, Weiwen and Tang, Ruiming and Wei, Zhewei and Zhang, Weinan and Yu, Yong},
  booktitle={Proceedings of the 48th International ACM SIGIR Conference on Research and Development in Information Retrieval},
  pages={1891--1901},
  year={2025}
}

@inproceedings{zheng2025enhancing,
  title={Enhancing embedding representation stability in recommendation systems with semantic id},
  author={Zheng, Carolina and Huang, Minhui and Pedchenko, Dmitrii and Rangadurai, Kaushik and Wang, Siyu and Xia, Fan and Nahum, Gaby and Lei, Jie and Yang, Yang and Liu, Tao and others},
  booktitle={Proceedings of the Nineteenth ACM Conference on Recommender Systems},
  pages={954--957},
  year={2025}
}

@inproceedings{freitag2017beam,
  title={Beam Search Strategies for Neural Machine Translation},
  author={Freitag, Markus and Al-Onaizan, Yaser},
  booktitle={Proceedings of the First Workshop on Neural Machine Translation},
  pages={56--60},
  year={2017}
}

@article{dao2022flashattention,
  title={Flashattention: Fast and memory-efficient exact attention with io-awareness},
  author={Dao, Tri and Fu, Dan and Ermon, Stefano and Rudra, Atri and R{\'e}, Christopher},
  journal={Advances in neural information processing systems},
  volume={35},
  pages={16344--16359},
  year={2022}
}

@inproceedings{daoflashattention,
  title={FlashAttention-2: Faster Attention with Better Parallelism and Work Partitioning},
  author={Dao, Tri},
  booktitle={The Twelfth International Conference on Learning Representations}
}

@inproceedings{lee2022autoregressive,
  title={Autoregressive image generation using residual quantization},
  author={Lee, Doyup and Kim, Chiheon and Kim, Saehoon and Cho, Minsu and Han, Wook-Shin},
  booktitle={Proceedings of the IEEE/CVF Conference on Computer Vision and Pattern Recognition},
  pages={11523--11532},
  year={2022}
}

@article{jegou2010product,
  title={Product quantization for nearest neighbor search},
  author={Jegou, Herve and Douze, Matthijs and Schmid, Cordelia},
  journal={IEEE transactions on pattern analysis and machine intelligence},
  volume={33},
  number={1},
  pages={117--128},
  year={2010},
  publisher={IEEE}
}

@article{kingma2014adam,
  title={Adam: A method for stochastic optimization},
  author={Kingma, Diederik P},
  journal={arXiv preprint arXiv:1412.6980},
  year={2014}
}

@article{dubey2024llama,
  title={The llama 3 herd of models},
  author={Dubey, Abhimanyu and Jauhri, Abhinav and Pandey, Abhinav and Kadian, Abhishek and Al-Dahle, Ahmad and Letman, Aiesha and Mathur, Akhil and Schelten, Alan and Yang, Amy and Fan, Angela and others},
  journal={arXiv e-prints},
  pages={arXiv--2407},
  year={2024}
}

@article{yang2025qwen3,
  title={Qwen3 technical report},
  author={Yang, An and Li, Anfeng and Yang, Baosong and Zhang, Beichen and Hui, Binyuan and Zheng, Bo and Yu, Bowen and Gao, Chang and Huang, Chengen and Lv, Chenxu and others},
  journal={arXiv preprint arXiv:2505.09388},
  year={2025}
}

@inproceedings{liu2024multi,
  title={Multi-behavior generative recommendation},
  author={Liu, Zihan and Hou, Yupeng and McAuley, Julian},
  booktitle={Proceedings of the 33rd ACM International Conference on Information and Knowledge Management},
  pages={1575--1585},
  year={2024}
}

@inproceedings{zhu2024cost,
  title={CoST: Contrastive Quantization based Semantic Tokenization for Generative Recommendation},
  author={Zhu, Jieming and Jin, Mengqun and Liu, Qijiong and Qiu, Zexuan and Dong, Zhenhua and Li, Xiu},
  booktitle={Proceedings of the 18th ACM Conference on Recommender Systems},
  pages={969--974},
  year={2024}
}

@article{liu2024end,
  title={End-to-End Learnable Item Tokenization for Generative Recommendation},
  author={Liu, Enze and Zheng, Bowen and Ling, Cheng and Hu, Lantao and Li, Han and Zhao, Wayne Xin},
  journal={arXiv preprint arXiv:2409.05546},
  year={2024}
}

@article{rafailov2023directDPO,
  title={Direct preference optimization: Your language model is secretly a reward model},
  author={Rafailov, Rafael and Sharma, Archit and Mitchell, Eric and Manning, Christopher D and Ermon, Stefano and Finn, Chelsea},
  journal={Advances in Neural Information Processing Systems},
  volume={36},
  pages={53728--53741},
  year={2023}
}

@inproceedings{hou2025generating,
  title={Generating long semantic IDs in parallel for recommendation},
  author={Hou, Yupeng and Li, Jiacheng and Shin, Ashley and Jeon, Jinsung and Santhanam, Abhishek and Shao, Wei and Hassani, Kaveh and Yao, Ning and McAuley, Julian},
  booktitle={Proceedings of the 31st ACM SIGKDD Conference on Knowledge Discovery and Data Mining V. 2},
  pages={956--966},
  year={2025}
}

@inproceedings{he2024rest,
  title={REST: Retrieval-Based Speculative Decoding},
  author={He, Zhenyu and Zhong, Zexuan and Cai, Tianle and Lee, Jason and He, Di},
  booktitle={Proceedings of the 2024 Conference of the North American Chapter of the Association for Computational Linguistics: Human Language Technologies (Volume 1: Long Papers)},
  pages={1582--1595},
  year={2024}
}

@article{STORE,
  title={STORE: Semantic Tokenization, Orthogonal Rotation and Efficient Attention for Scaling Up Ranking Models},
  author={Xu, Yi and Fan, Chaofan and Hu, Jinxin and Zhang, Yu and Xiaoyi, Zeng and Zhang, Jing},
  journal={arXiv preprint arXiv:2511.18805},
  year={2025}
}

@article{zhang2025onetrans,
  title={OneTrans: Unified Feature Interaction and Sequence Modeling with One Transformer in Industrial Recommender},
  author={Zhang, Zhaoqi and Pei, Haolei and Guo, Jun and Wang, Tianyu and Feng, Yufei and Sun, Hui and Liu, Shaowei and Sun, Aixin},
  journal={arXiv preprint arXiv:2510.26104},
  year={2025}
}

@inproceedings{zhu2025rankmixer,
  title={Rankmixer: Scaling up ranking models in industrial recommenders},
  author={Zhu, Jie and Fan, Zhifang and Zhu, Xiaoxie and Jiang, Yuchen and Wang, Hangyu and Han, Xintian and Ding, Haoran and Wang, Xinmin and Zhao, Wenlin and Gong, Zhen and others},
  booktitle={Proceedings of the 34th ACM International Conference on Information and Knowledge Management},
  pages={6309--6316},
  year={2025}
}

@article{fu2025forge,
  title={Forge: Forming semantic identifiers for generative retrieval in industrial datasets},
  author={Fu, Kairui and Zhang, Tao and Xiao, Shuwen and Wang, Ziyang and Zhang, Xinming and Zhang, Chenchi and Yan, Yuliang and Zheng, Junjun and Li, Yu and Chen, Zhihong and others},
  journal={arXiv preprint arXiv:2509.20904},
  year={2025}
}

@inproceedings{ju2025generative,
  title={Generative Recommendation with Semantic IDs: A Practitioner's Handbook},
  author={Ju, Clark Mingxuan and Collins, Liam and Neves, Leonardo and Kumar, Bhuvesh and Wang, Louis Yufeng and Zhao, Tong and Shah, Neil},
  booktitle={Proceedings of the 34th ACM International Conference on Information and Knowledge Management},
  pages={6420--6425},
  year={2025}
}

@article{yang2024unifying,
  title={Unifying generative and dense retrieval for sequential recommendation},
  author={Yang, Liu and Paischer, Fabian and Hassani, Kaveh and Li, Jiacheng and Shao, Shuai and Li, Zhang Gabriel and He, Yun and Feng, Xue and Noorshams, Nima and Park, Sem and others},
  journal={arXiv preprint arXiv:2411.18814},
  year={2024}
}

@inproceedings{zhai2024actions,
  title={Actions Speak Louder than Words: Trillion-Parameter Sequential Transducers for Generative Recommendations},
  author={Zhai, Jiaqi and Liao, Lucy and Liu, Xing and Wang, Yueming and Li, Rui and Cao, Xuan and Gao, Leon and Gong, Zhaojie and Gu, Fangda and He, Jiayuan and others},
  booktitle={International Conference on Machine Learning},
  pages={58484--58509},
  year={2024},
  organization={PMLR}
}

@article{li2025survey,
  title={A survey of generative recommendation from a tri-decoupled perspective: Tokenization, architecture, and optimization},
  author={Li, Xiaopeng and Chen, Bo and She, Junda and Cao, Shiteng and Wang, You and Jia, Qinlin and He, Haiying and Zhou, Zheli and Liu, Zhao and Liu, Ji and others},
  year={2025},
  publisher={Preprints}
}

@inproceedings{gao2025generative,
  title={Generative auto-bidding with value-guided explorations},
  author={Gao, Jingtong and Li, Yewen and Mao, Shuai and Jiang, Peng and Jiang, Nan and Wang, Yejing and Cai, Qingpeng and Pan, Fei and Jiang, Peng and Gai, Kun and others},
  booktitle={Proceedings of the 48th International ACM SIGIR Conference on Research and Development in Information Retrieval},
  pages={244--254},
  year={2025}
}

@article{wang2025gflowgr,
  title={GFlowGR: Fine-tuning Generative Recommendation Frameworks with Generative Flow Networks},
  author={Wang, Yejing and Zhou, Shengyu and Lu, Jinyu and Liu, Qidong and Li, Xinhang and Zhang, Wenlin and Li, Feng and Wang, Pengjie and Xu, Jian and Zheng, Bo and others},
  journal={arXiv preprint arXiv:2506.16114},
  year={2025}
}

@article{fu2025unified,
  title={A unified framework for multi-domain ctr prediction via large language models},
  author={Fu, Zichuan and Li, Xiangyang and Wu, Chuhan and Wang, Yichao and Dong, Kuicai and Zhao, Xiangyu and Zhao, Mengchen and Guo, Huifeng and Tang, Ruiming},
  journal={ACM Transactions on Information Systems},
  volume={43},
  number={5},
  pages={1--33},
  year={2025},
  publisher={ACM New York, NY}
}

@inproceedings{zhang2025notellm,
  title={Notellm-2: Multimodal large representation models for recommendation},
  author={Zhang, Chao and Zhang, Haoxin and Wu, Shiwei and Wu, Di and Xu, Tong and Zhao, Xiangyu and Gao, Yan and Hu, Yao and Chen, Enhong},
  booktitle={Proceedings of the 31st ACM SIGKDD Conference on Knowledge Discovery and Data Mining V. 1},
  pages={2815--2826},
  year={2025}
}

@inproceedings{wang2025put,
  title={Put Teacher in Student's Shoes: Cross-Distillation for Ultra-compact Model Compression Framework},
  author={Wang, Maolin and Chu, Jun and Xie, Sicong and Zang, Xiaoling and Zhao, Yao and Zhong, Wenliang and Zhao, Xiangyu},
  booktitle={Proceedings of the 31st ACM SIGKDD Conference on Knowledge Discovery and Data Mining V. 2},
  pages={4975--4985},
  year={2025}
}

% \appendix
% \input{7Appendix}

\end{sloppypar}
\end{document}